\documentclass[runningheads]{llncs}
\usepackage{graphicx}
\usepackage[colorlinks,
            linkcolor=red,
            anchorcolor=blue,
            citecolor=green
            ]{hyperref}
\usepackage{amsfonts}
\usepackage{bm}
\usepackage{amsmath}
\usepackage{autobreak}
\usepackage{xcolor}
\usepackage{multirow}
\usepackage{marvosym}
\begin{document}
\title{Interpretable Spatio-Temporal Embedding for Brain Structural-Effective Network with Ordinary Differential Equation}
% Brain Dynamics Embeeding via Ordinary Differential Equation

\titlerunning{STE-ODE for Brain Network Learning}

\author{Haoteng Tang\inst{1}\textsuperscript{\Letter}
\and Guodong Liu\inst{2}  
\and Siyuan Dai\inst{3} 
\and Kai Ye\inst{3} 
\and Kun Zhao\inst{3} 
\and Wenlu Wang\inst{4} 
\and Carl Yang\inst{5} 
\and Lifang He\inst{6}
\and Alex Leow\inst{7}
\and Paul Thompson\inst{8}
\and Heng Huang\inst{2} 
\and Liang Zhan\inst{3} \textsuperscript{\Letter}
}
\authorrunning{H. Tang et al.}
\institute{University of Texas Rio Grande Valley, Edinburg, TX, 78539, USA \\
\email{haoteng.tang@utrgv.edu} \and
University of Maryland, College Park, MD, 20742, USA \and
University of Pittsburgh, Pittsburgh, PA, 15260, USA \\
\email{liang.zhan@pitt.edu}\and
Texas A\&M University - Corpus Christi, Corpus Christi, TX, 78412, USA \and
Emory University, Atlanta, GA, 30322, USA \and
Lehigh University, Bethlehem, PA, 18015, USA \and
University of Illinois at Chicago, Chicago, IL, 60612, USA \and
University of Southern California, Los Angeles, CA, 90032, USA
}

\maketitle              % typeset the header of the contribution
\begin{abstract}
% MRI-based modeling of brain networks has been widely used to understand brain functionalities and structures, as well as factors that affect them, such as brain diseases and development. 
% Graph representation learning on MRI-derived brain networks may facilitate the discovery of novel biomarkers for clinical phenotypes and neurodegenerative diseases.
% Most of the current brain network learning methods focus on analyzing functional brain networks derived from synchronous BOLD signals, which may not capture the directional influences among brain regions. 
% Meanwhile, few of the current studies focus on modeling the brain temporal functional dynamics. 
% In this study, we combine the analysis on structural networks as well as temporal effective networks derived by fMRI BOLD signals. 
% Particularly, we propose an interpretable graph learning framework, \textcolor{red}{Spatio-Temporal ODE (STE-ODE)}, with directed node embedding layers, which captures structural-effective network representations by solving an
% ordinary differential equation that models the brain spatial-temporal dynamics.
% We validate our framework on clinical phenotype and neurodegenerative disease prediction tasks using two independent, publicly available datasets (HCP and OASIS). 
% The experimental results clearly demonstrate the advantages of our model compared to several state-of-the-art methods. 
The MRI-derived brain network serves as a pivotal instrument in elucidating both the structural and functional aspects of the brain, encompassing the ramifications of diseases and developmental processes.
However, prevailing methodologies, often focusing on synchronous BOLD signals from functional MRI (fMRI), may not capture directional influences among brain regions and rarely tackle temporal functional dynamics. 
In this study, we first construct the brain-effective network via the dynamic causal model. 
%Then, we design a directed graph embedding layer for effective network embedding under the constraints of the diffusion MRI-derived structural network.
% In this study, we introduce a novel approach to generate temporal effective networks from fMRI BOLD signals constrained by brain structure networks. 
Subsequently, we introduce an interpretable graph learning framework termed Spatio-Temporal Embedding ODE (STE-ODE). 
This framework incorporates specifically designed directed node embedding layers, aiming at capturing the dynamic interplay between structural and effective networks via an ordinary differential equation (ODE) model, which characterizes spatial-temporal brain dynamics.
%Then, we present an interpretable graph learning framework, Spatio-Temporal Embedding ODE (STE-ODE), featuring designed directed node embedding layers that capture the dynamic interplay of structural and effective networks through an ordinary differential equation (ODE) modeling spatial-temporal brain dynamics.
Our framework is validated on several clinical phenotype prediction tasks using two independent publicly available datasets (HCP and OASIS). 
The experimental results clearly demonstrate the advantages of our model compared to several state-of-the-art methods.

\end{abstract}

\begin{small}
\noindent\textbf{\textcolor{blue}{Disclaimer.}}
\textcolor{blue}{This preprint is an early version of the paper that has been accepted by MICCAI 2024. This version has not yet undergone final formatting and proofreading. The final version will be published in the MICCAI conference proceedings.} 
\end{small}
\section{Introduction}
% Neuroimaging, brain networks, brain disease and clinical phenotypes. 
% Different kind of brain network -- structural, dynamic=(functional, effective). Difference between functional and effective --> causality with time lag. 
% How to model these series of brain effective networks (dynamic)? -- Effective Graph ODE
% Interpretable issue. How to identify the changes of significant effective weights during thie timeline? How these weightes changes?

% Neuroimaging techniques (e.g., Magnetic Resonance Imaging (MRI)) offer a non-invasive method to visualize brain structure and function in vivo, which have advanced our understanding of human brain organizations. 
% Recent years have witnessed great progress in applying network science methods to study MRI-derived brain networks, to discover novel biomarkers for clinical phenotypes or neurodegenerative diseases (e.g., Alzheimer’s disease or AD).
% Brain networks, constructed by analyzing patterns of structural or functional connectivity obtained from MRI data, represent a complex 3D brain graph model of neural connections to enable researchers to study the interaction across different brain regions-of-interest (ROIs).
Neuroimaging techniques, such as Magnetic Resonance Imaging (MRI), have significantly advanced our understanding of the brain by providing a non-invasive way to explore its anatomical structures and functions. 
Recent advances in network science have allowed for the analysis of MRI-derived brain networks, revealing new biomarkers for diseases such as Alzheimer's and enabling the study of complex neural interactions across different brain regions \cite{bullmore2009complex}.

% Different MRI techniques can be utilized to reconstruct brain networks providing different unique insights of brain organizations or dynamics. 
% Particularly, diffusion MRI (or dMRI) measures the diffusion process of water molecules in brain tissues to map white matter tractography in the brain. 
% The dMRI-derived brain networks map anatomical connections among different brain ROIs to construct the brain structural connectivity, revealing how different brain ROIs are physically connected through neural pathways.
% Beyond brain structures, functional MRI (or fMRI) provides the waves of blood-oxygen level-dependent (or BOLD) signals to measure the brain dynamic activities.
Different MRI techniques reveal diverse aspects of brain organization and dynamics. For example, diffusion MRI (dMRI) maps white matter connections by tracking water molecule diffusion, showing how brain regions are structurally linked. 
Functional MRI (fMRI), on the other hand, utilizes blood-oxygen level-dependent (BOLD) signals to monitor brain activity, offering insights into functional brain dynamics.
% Many recent studies derive functional brain networks from fMRI BOLD signals by estimating temporal correlations in BOLD signals across different brain regions (e.g., Pearson Correlation) to identify patterns of functional connectivity.
% However, the functional networks model the interaction between two brain regions by utilizing the BOLD signals at the same time interval, which may not capture the directional influence between these regions over time. 
% This time-related directional influence among brain regions is the so called brain region causality.
Recent research utilizing fMRI BOLD signals to delineate functional brain networks has made significant strides in identifying patterns of connectivity through temporal correlations (e.g., Pearson correlation) across different brain regions. 
These studies highlight the utility of fMRI in mapping the intricate web of neural interactions, presenting the brain's complex connectivity patterns \cite{shinn2023functional,guo2023investigating}. 
However, traditional methods primarily focus on synchronous BOLD signals, which may overlook the nuanced directional influences (e.g., causality) between brain regions over time.
% To model the causality among brain regions, we utilize time-lagged BOLD signals to build up effective connectivity based on Dynamic Causal Modeling (DCM) architecture \cite{friston2003dynamic}, yielding directed effective brain networks \cite{tang2023comprehensive}.  
% Effective connectivity embodies the causal influence that the functional activity of a source brain region exerts over the activity of a target brain region over time\cite{chuang2023brain}.
To capture the directional influences among brain regions, we employ Dynamic Causal Modeling (DCM) \cite{chuang2023brain,friston2003dynamic} with time-lagged BOLD signals to construct temporal effective connectivity networks. %chuang2023brain
The temporal effective networks represent the dynamic causal relationships where the activity of one brain region influences another over time.
% In recent years, Graph neural networks (GNNs) \cite{kipf2016semi} have gained enormous attention and achieved great progress in brain network studies, particularly for brain structural and functional network representations \cite{amodeo2022unified,tang2023signed}.
% However, very few graph learning methods are proposed for effective network learning \cite{chen2023fe}.
% It is notable that the effective brain network for each subject is a series of directed brain graphs across time, where the effective connectivity changes over time. 
% This may result in two different challenges when we build up GNNs on effective brain networks.

In recent years, Graph Neural Networks (GNNs) \cite{kipf2016semi} have become increasingly prominent in brain network studies, showing significant advancements in mining brain structural and functional networks \cite{amodeo2022unified,wu2021federatediccad,wu2021federatedmiccai,jia2020personalized,liu2020sparse,tang2022contrastive,tang2022hierarchical,tang2023comprehensive,tang2023signed,ye2023bidirectional}. 
Despite this progress, a scarcity of graph learning methods is designed for dynamic effective network learning \cite{chen2023fe}. 
The dynamic effective brain networks are a series of time-evolving directed graphs, which may present two challenges when we build up GNNs on these networks.  
First, existing GNNs focused on embedding nodes in undirected graphs, which may not effectively handle directed graph embeddings. 
Effective brain networks feature pairs of brain regions connected by directed edges with different weights, where the edge direction and weight represent the causal sequence and its magnitude, respectively. 
To address this, we propose a directed graph encoder specifically designed for capturing these causal sequences in brain node embedding.
Furthermore, the dynamic effective brain network consists of temporal sequences of brain graphs, with changing connectivity over time. 
Thus, current GNNs need to be adapted to capture both spatial and temporal dynamics of the brain. 
Recent efforts in dynamic graph learning include approaches such as recurrent graph neural network \cite{demirbilek2021recurrent}, graph temporal attention network \cite{li2023stgate}, and graph transformer \cite{zhao2022revealing}. 
In this study, we tackle the brain spatial-temporal dynamics with an ordinary differential equation (ODE) model.
Particularly, we introduce a graph learning framework, Spatio-Temporal Embedding ODE (STE-ODE), designed to simultaneously solve an Ordinary Differential Equation (ODE) and embed brain networks, capturing both their structural and functional properties. 
The framework's unique approach ensures that the training process yields brain network embeddings that are, in essence, solutions to the ODE, thereby intertwining the learning model with the ODE resolution.
% We present a graph learning framework, dubbed Spatio-Temporal Embedding ODE (STE-ODE), which simultaneously solves this ODE while embedding the structural and effective brain networks, where the optimal brain network embedding is the solution of this ODE. 
These embedded graph representations are then leveraged for different clinical predictions, such as brain disease classifications. 
Beyond prediction tasks, our study aims to identify most significant connectomes related to various clinical phenotypes and neurodegenerative diseases, tracking their changes over time for different tasks. 
To this end, we develop an interpretable toolkit within our directed node embedding layer. 
This toolkit focuses on pinpointing the top $K$ edges with significant temporal changes, marking them as potential biomarkers for distinct phenotypes. This method directly connects dynamic brain network changes to specific biological traits, enhancing our comprehension of the mechanisms tied to different phenotypes.
Our contributions can be summarized as follows. (1) We design a directed graph embedding layer tailored for encoding effective network under the constrains of its structural counterpart. 
(2) We present a learning framework with the directed graph embedding layer, referred to as STE-ODE, which captures temporal effective network representations by solving an ordinary differential equation that models the brain spatial-temporal dynamics. 
(3) We develop a toolkit to enhance the interpretability of our framework, which enables the identification of the most significant connectome changes, marking them as potential biomarkers for different clinical phenotypes.

\begin{figure}[t]
\centering
\includegraphics[width=1.0\textwidth]{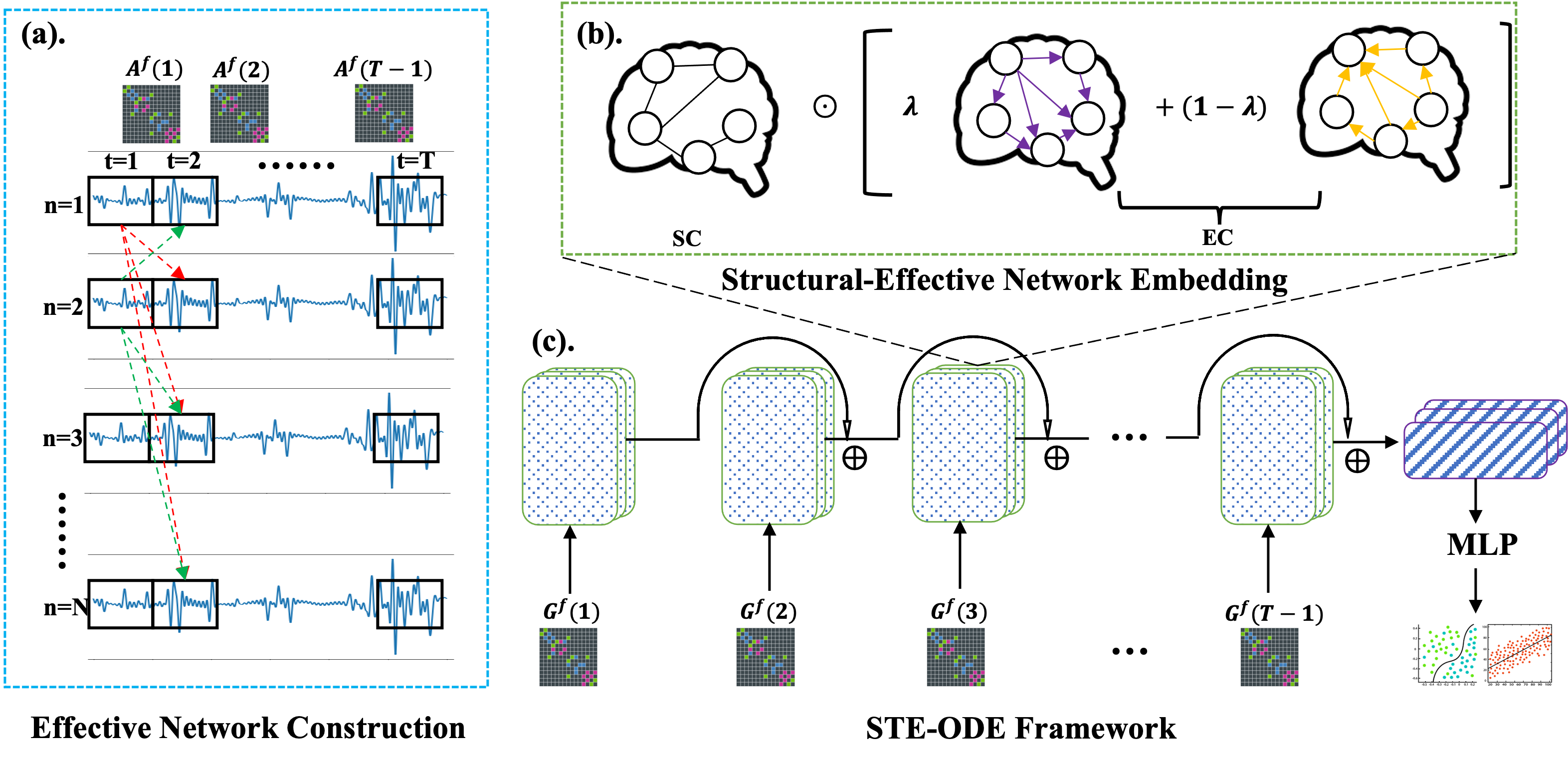}
\caption{(a) describes the construction of brain effective networks from the BOLD signals. (b) is the directed graph embedding layer for structural and effective networks. (c) presents the STE-ODE framework for different clinical prediction tasks.}
\label{framework}
\end{figure}
\vspace{-1em}
\section{Methodology}
% prelimiary
% effective network construction
% directed graph node encoder with interpretable toolkit
% GNN framework for achieve graph ODE
We first introduce our method for constructing directed effective networks through the dynamic causal model (DCM). 
Additionally, we propose our interpretable directed graph node embedding layer, which is tailored to encode both directed effective networks and their structural counterparts. 
Subsequently, we detail our comprehensive spatio-temporal framework with the directed graph embedding layer for downstream tasks. 
This framework involves solving an ordinary differential equation that captures the spatial-temporal dynamics of the brain.
% We first introduce our method for constructing directed effective networks using the dynamic causal model (DCM).
% Additionally, we introduce our proposed directed graph node embedding layer, designed to encode directed effective networks as well as the structural counterparts.
% Following this, we detail our comprehensive spatio-temporal framework with the directed graph embedding layer for downstream tasks, which is achieved by solving an ordinary differential equation that models the spatial-temporal dynamics of the brain.

\subsection{Preliminaries}
A brain network is a weighted graph $G=\{V, E\}=(A,X)$ with $N$ nodes, where $V=\{v_i\}_{i=1}^{N}$ is the set of graph nodes representing brain regions, and $E=\{e_{i,j}\}$ is the edge set. 
$X \in \mathbb{R}^{N \times c}$ is the node feature matrix where $x_{i} \in \mathbb{R}^{1 \times c}$ is the $i-$th row of $X$ representing the node feature of $v_{i}$.
$A \in \mathbb{R}^{N \times N}$ is the adjacency matrix where $a_{i,j} \in \mathbb{R}$ represents the weights of the edge between $v_{i}$ and $v_{j}$. 
A brain structural network, denoted as $G^{s}$, is an undirected graph, where $e^{s}_{i,j}=e^{s}_{j,i} \geq 0$. 
In stead, a brain effective network, denoted as $G^{f}$, is a directed graph, where $e^{f}_{i,j} \neq e^{f}_{j,i} \in \mathbb{R}$.
The sign of $e^{f}_{i,j}$ indicates the causal sequence between $v_{i}$ and $v_{j}$, where $e^{f}_{i,j}>0$ signifies the causal effect on $v_{j}$ induced by $v_{i}$, vice versa. Additionally, we denote the blood-oxygen-level-dependent (BOLD) signal obtained from fMRI as $B \in \mathbb{R}^{N \times b}$.

\subsection{Construction of Brain Effective Network}
%sanchez2019estimating
We employ fMRI BOLD signals to construct brain effective networks using the dynamic causal modeling (DCM) approach \cite{sanchez2019estimating,smith2011network}. Each brain region serves as a graph node embedded within the brain effective network, while the temporal dynamic effective connectivity comprises the edge set. Given the fMRI BOLD signals, the dynamic adjacency matrix $A^{f}(t)$ can be modeled as follows:
\begin{equation}
    \frac{dB(t)}{dt} = \alpha A^{f}(t)B(t) + Cu(t)
\label{adj-ode}
\end{equation}
$Cu(t)$ represents the term governing the influence of external neuronal inputs $u(t)$ on the dynamics of $A^{f}$. In this work, $Cu(t)=0$ as we concentrate on resting-state fMRI studies. 
The parameter $\alpha$ serves as a constant regulating the neuronal lag among brain nodes. Consequently, we can derive the expression of $A^{f}$ as follows:
\begin{equation}
    A^{f}(t) = \frac{1}{\alpha B(t)}\frac{dB(t)}{dt}
    \label{Af-continuous}
\end{equation}
We construct the effective connectivity by deriving the discrete expression of the Eq. (\ref{Af-continuous}):
\begin{equation}
    A^{f}(t) = \frac{1}{\alpha B(t)}\frac{B(t+1)-B(t)}{t+1-t}
    = \frac{1}{\alpha}(\frac{B(t+1)}{B(t)} - 1)
\end{equation}
We define the connectivity between brain node $v^{f}_{i}$ and $v^{f}_{j}$ at timepoint $t$ as follows, with $\beta=\frac{1}{\alpha} \in [0,1]$:
\begin{equation}
    A^{f}_{i,j}(t) = \beta (\frac{B_{j}(t+1)}{B_{i}(t)} - 1),
\end{equation}
where $B_{i}$ is the BOLD signal of $v_{i}$. 
The process of constructing brain effective networks is illustrated in Fig. \ref{framework}(a).

\subsection{Interpretable Structural-Effective Network Embedding}
Given a directed effective network $G^{f}=(A^{f},X^{f})$, we first perform asymmetric Laplacian normalization on its adjacency matrix. The normalized adjacency matrix can be represented as:
\begin{equation}
    \widetilde{A}^{f} = D_{\text{in}}^{-\frac{1}{2}} A^{f} D_{\text{out}}^{-\frac{1}{2}},
\end{equation}
where $D_{\text{in}}$ and $D_{\text{out}}$ are in-degree and out-degree of the adjacency matrix, respectively. 
Then, our node embedding layer for the structural-effective network can be formulated as a function $\mathcal{F}_{\mathcal{G}}$:
\begin{eqnarray}
    \overline{Z} &=& \mathcal{F}_{\mathcal{G}}(\widetilde{A}^{s},\widetilde{A}^{f},X^{f}; W,\gamma,\lambda) \nonumber \\
    &=& \sigma(\gamma \odot \widetilde{A}^{s} \odot [\lambda \widetilde{A}^{f} + (1-\lambda)\widetilde{A}^{f\top}] X^{f} W), 
    \label{node-embedding-layer}
\end{eqnarray}
where $\widetilde{A}^{s}$ represents the Laplacian-normalized adjacency matrix of the brain structural network \cite{kipf2016semi}.
The brain structural network serves as spatial information to constrain the temporal function dynamics, under the assumption that two brain regions are functionally interconnected as long as they are structurally connected \cite{stam2016relation}. %zhang2020deep
$\sigma(\cdot)$ is a nonlinear activation function, such as \textit{ReLU}.
$\lambda \in [0,1]$ is a parameter that balances the information flow into and out of each brain node.
$W$ represents trainable parameters for brain node embedding.
$\gamma \in \mathbb{R}^{N \times N}$ are trainable parameters used for model interpretability, enabling edge weights to adapt themselves for different prediction targets.
During the model validation stage, we utilize self-adapted edge weights to track the most important connectomes for various prediction tasks.
% It is worth noting that in the initial embedding layer, the input $Z^{f}$ is initialized as the node feature matrix $X^{f}$. 
The brain node embedding layer is depicted in Fig. \ref{framework}(b).
\vspace{-1em}
\subsection{Spatio-Temporal Embedding with ODE}
% \textbf{Spatio-Temporal Representation Learning with ODE.}
Given a series of temporal effective networks (i.e., $G^{f}(t), t \in [0,T]$), their dynamic embeddings can be modeled using the following ordinary differential equation:
\begin{equation}
    \mathcal{F}_{\mathcal{G}}(G^{f}(t+\triangle t),\mathbf{\Theta})
  = \mathcal{F}_{\mathcal{G}}(G^{f}(t),\mathbf{\Theta}) + \int_{t}^{t+\triangle t}     \mathcal{F}_{\mathcal{G}}(G^{f}(\tau),\mathbf{\Theta})d \tau,
   \label{brain-ode-continuous}
\end{equation}
where $\mathbf{\Theta}$ is the parameter sets (i.e., $\mathbf{\Theta}=\{W, \gamma, \lambda\}$) of the embedding function.
We can approximate the Eq. \ref{brain-ode-continuous} into the discrete expression with our proposed node embedding layer (see Eq. \ref{node-embedding-layer}) as:
\begin{align}
    \overline{Z}(t+1)
  = \overline{Z}(t) 
  + \sigma(\gamma \widetilde{A}^{s} \odot [\lambda \widetilde{A}^{f}(t+1) + (1-\lambda)\widetilde{A}^{f\top}(t+1)] X(t+1) W).
\end{align}
We unfold the temporal brain network embedding into an residual graph learning framework. 
In this framework, each embedding layer processes the dynamic effective network at $G^{f}(t+1)$, while the previous dynamic network embedding (i.e., $\overline{Z}(t)$) is treated as a residual term.
\vspace{-1em}
\subsection{STE-ODE Framework for Brain Network Predictions}
The proposed STE-ODE framework, incorporating the spatio-temporal embedding model, is depicted in Fig. \ref{framework}(c).
Assuming we have obtained the last node embedding (i.e., $\overline{Z}(T)$), we employ an average global pooling layer ($Z_{G}=\frac{1}{N}\sum_{i=1}^{N} \\ \overline{Z}_{i}(T)$) to extract the entire graph representation. Subsequently, a fully connected neural network (such as a Multilayer Perceptron or MLP) is employed to generate the final classification or regression output (i.e., $\hat{y} = MLP(Z_{G})$).
For the classification task, we utilize the negative log likelihood loss function, where $\mathcal{L} = NLL\_Loss(\hat{y}, y)$.
For the regression task, we use the $L_{2}$ loss function, where $\mathcal{L} = L_{2}Loss(\hat{y}, y)$.

\begin{table}[t]
\centering
\caption{Classification accuracy and F1-scores, along with their standard deviations under 5-fold cross-validation. The best results are highlighted in \textcolor{red}{red}.}
\label{classification}
\setlength\tabcolsep{2.6pt}
\scalebox{0.95}{
\begin{tabular}{c|ll|llll}
\hline
\multirow{3}{*}{Method} & \multicolumn{2}{c|}{HCP}                                   & \multicolumn{4}{c}{OASIS}                                                                                                       \\ \cline{2-7} 
                        & \multicolumn{2}{c|}{Gender}                                & \multicolumn{2}{c|}{Disease}                                        & \multicolumn{2}{c}{$\epsilon 4$}                                \\ \cline{2-7} 
                        & \multicolumn{1}{c|}{Acc.}        & \multicolumn{1}{c|}{F1} & \multicolumn{1}{c|}{Acc.}        & \multicolumn{1}{c|}{F1}          & \multicolumn{1}{c|}{Acc.}        & \multicolumn{1}{c}{F1} \\ \hline
SVM                     & \multicolumn{1}{l|}{59.25$\pm$1.39} & 60.85$\pm$2.29$\pm$             & \multicolumn{1}{l|}{57.72$\pm$0.98} & \multicolumn{1}{l|}{56.58$\pm$1.93} & \multicolumn{1}{l|}{58.09$\pm$2.37} & 59.83$\pm$0.99            \\
GCN                     & \multicolumn{1}{l|}{68.83$\pm$1.48} & 67.48$\pm$2.32             & \multicolumn{1}{l|}{64.64$\pm$1.05} & \multicolumn{1}{l|}{66.58$\pm$2.12} & \multicolumn{1}{l|}{65.56$\pm$1.51} & 64.28$\pm$1.11            \\
DiffPool                & \multicolumn{1}{l|}{73.25$\pm$0.71} & 70.43$\pm$1.87             & \multicolumn{1}{l|}{71.67$\pm$0.83} & \multicolumn{1}{l|}{69.58$\pm$1.75} & \multicolumn{1}{l|}{69.04$\pm$2.52} & 70.42$\pm$0.87            \\ \hline
LSTM                    & \multicolumn{1}{l|}{70.95$\pm$1.09} & 72.37$\pm$2.16             & \multicolumn{1}{l|}{68.22$\pm$2.04} & \multicolumn{1}{l|}{68.90$\pm$0.74} & \multicolumn{1}{l|}{69.33$\pm$1.88} & 67.31$\pm$2.65            \\
ST-GCN                  & \multicolumn{1}{l|}{78.44$\pm$0.86} & 76.15$\pm$1.17             & \multicolumn{1}{l|}{76.26$\pm$0.98} & \multicolumn{1}{l|}{77.02$\pm$1.47} & \multicolumn{1}{l|}{77.20$\pm$1.79} & 78.14$\pm$1.35            \\
FE-STGNN                & \multicolumn{1}{l|}{81.04$\pm$0.39} & 81.75$\pm$1.26             & \multicolumn{1}{l|}{79.92$\pm$0.73} & \multicolumn{1}{l|}{79.39$\pm$1.15} & \multicolumn{1}{l|}{78.98$\pm$0.92} & 80.06$\pm$0.85            \\ \hline
Ours w/o SC             & \multicolumn{1}{l|}{80.66$\pm$2.02} & 80.77$\pm$0.63             & \multicolumn{1}{l|}{\textcolor{red}{80.59$\pm$1.71}} & \multicolumn{1}{l|}{81.05$\pm$1.20} & \multicolumn{1}{l|}{78.42$\pm$1.07} & 78.59$\pm$1.63            \\
Ours                    & \multicolumn{1}{l|}{\textcolor{red}{82.12$\pm$1.17}} & \textcolor{red}{83.97$\pm$0.96}             & \multicolumn{1}{l|}{80.01$\pm$1.26} & \multicolumn{1}{l|}{\textcolor{red}{81.31$\pm$1.37}} & \multicolumn{1}{l|}{\textcolor{red}{81.35$\pm$0.86}} & \textcolor{red}{80.92$\pm$1.03}            \\ \hline
\end{tabular}}
\end{table}

\begin{table}[t]
\centering
\caption{Regression mean absolute values with their \emph{std} under 5-fold cross-validation. The best results are highlighted in \textcolor{red}{red}.}
\label{regression}
\begin{tabular}{c|ccc|c}
\hline
\multirow{2}{*}{Method} & \multicolumn{3}{c|}{HCP}                                                  & OASIS    \\ \cline{2-5} 
                        & \multicolumn{1}{c|}{MMSE}     & \multicolumn{1}{c|}{DSM-Depr} & DSM-Antis & MMSE     \\ \hline
SVM                     & \multicolumn{1}{c|}{4.06$\pm$0.33} & \multicolumn{1}{c|}{4.66$\pm$0.79} & 3.43$\pm$0.59  & 3.91$\pm$0.24 \\
GCN                     & \multicolumn{1}{c|}{3.16$\pm$0.43} & \multicolumn{1}{c|}{3.62$\pm$0.98} & 3.41$\pm$0.37  & 3.70$\pm$1.06 \\
DiffPool                & \multicolumn{1}{c|}{2.82$\pm$0.93} & \multicolumn{1}{c|}{3.23$\pm$0.54} & 2.09$\pm$0.56  & 2.48$\pm$0.90 \\ \hline
LSTM                    & \multicolumn{1}{c|}{2.74$\pm$0.91} & \multicolumn{1}{c|}{2.37$\pm$0.61} & 1.91$\pm$0.47  & 1.88$\pm$0.51 \\
ST-GCN                  & \multicolumn{1}{c|}{1.97$\pm$0.84} & \multicolumn{1}{c|}{1.35$\pm$0.17} & 1.24$\pm$0.33  & 1.19$\pm$0.23 \\
FE-STGNN                & \multicolumn{1}{c|}{0.73$\pm$0.29} & \multicolumn{1}{c|}{1.19$\pm$0.14} & 1.08$\pm$0.06  & 0.96$\pm$0.15 \\ \hline
Ours w/o SC             & \multicolumn{1}{c|}{0.93$\pm$0.44} & \multicolumn{1}{c|}{1.24$\pm$0.32} & 1.19$\pm$0.24  & 1.08$\pm$0.33 \\
Ours                    & \multicolumn{1}{c|}{\textcolor{red}{0.62$\pm$0.23}} & \multicolumn{1}{c|}{\textcolor{red}{1.08$\pm$0.45}} & \textcolor{red}{0.92$\pm$0.79}  & \textcolor{red}{0.76$\pm$0.17} \\ \hline
\end{tabular}
\end{table}

\section{Experiments}
\subsection{Dataset Description and Preprocessing}
Two publicly available datasets were used to evaluate our framework.
The first includes data from $1206$ young healthy subjects (mean age $28.19 \pm 7.15$, $657$ women) from the Human Connectome Project \cite{van2013wu} (HCP). 
The second includes 1326 subjects (mean age $=70.42 \pm 8.95$, $738$ women) from the Open Access Series of Imaging Studies (OASIS) dataset \cite{lamontagne2019oasis}. 
Details of each dataset can be found on their official websites. 
% Detailed information on each dataset can be accessed through their respective official websites \footnote{\url{https://www.humanconnectome.org/study/hcp-young-adult/data-releases}\\ \url{https://www.oasis-brains.org}}. 
The preprocessing of functional BOLD signals and the reconstruction of structural networks were conducted using CONN \cite{whitfield2012conn} and FSL Probtrackx\cite{jenkinson2012fsl}, respectively. 
For the HCP data, both structural and effective networks have a dimension of $82 \times 82$ based on 82 ROIs defined using FreeSurfer (V6.0)~\cite{fischl2012freesurfer}. 
For the OASIS data, both networks have a dimension of $132 \times 132$ based on the Harvard-Oxford Atlas and AAL Atlas. 
% In the case of the HCP dataset, both structural and effective networks were structured into an $82 \times 82$ matrix, reflecting $82$ Regions of Interest (ROIs) as delineated by FreeSurfer (V6.0) \cite{fischl2012freesurfer}. 
% While for the OASIS dataset, the networks were arranged into a $132 \times 132$ matrix, informed by the Harvard-Oxford Atlas and AAL Atlas. 
This intentional variation in network resolutions for the HCP and OASIS datasets served to examine whether the dimension of the network or the choice of atlas influences the efficacy of our newly developed framework.

\subsection{Implementation Details and Experimental Setup}
\begin{figure}[htpb]
\centering
\includegraphics[width=1.0\textwidth]{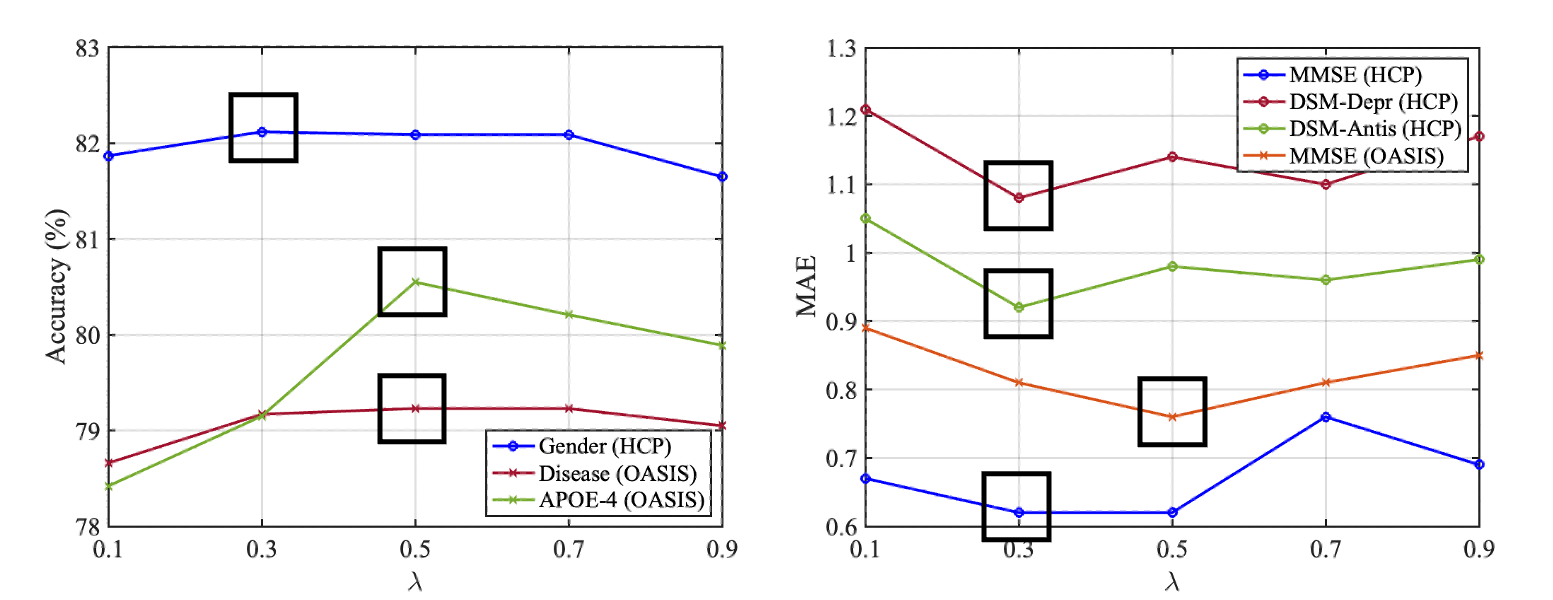}
\vspace{-26pt}
\caption{$\lambda$ parameter analysis. The optimal points are encompassed by black boxes.}
\label{parameter}
\end{figure}

\textbf{\textit{Implementation Details.}}
We divided the BOLD signal $B$ into $T=5$ time segments and calculated the mean value of the points within each segment to construct $4$ effective networks.
The edge weights of both the effective networks and structural networks were normalized to the intervals $[-1, 1]$ and $[0, 1]$, respectively. 
Node features were initialized by sampling from a standard Gaussian distribution with feature dimensions set to $16$.
Each dataset was randomly partitioned into $5$ disjoint sets for $5$-fold cross-validation in subsequent experiments.
The Adam optimizer was utilized to train the model with a batch size of $128$. 
The initial learning rate was set to $0.001$ and decayed by $(1-\frac{\text{current epoch}}{\text{max epoch}})^{0.9}$.
We also regularized the training with an $L_{2}$ weight decay of $1e^{-5}$.
% The model is trained using the Adam optimizer with a batch size of $128$. 
% The initial learning rate is set to $0.001$ and decayed by $(1-\frac{current \, epoch}{max \, epoch})^{0.9}$.
% We also regularize the training with an $L_{2}$ weight decay of $1e^{-5}$.
% We stop the training if the validation loss does not improve for $100$ epochs in an epoch termination condition as was done in \cite{lee2019self,shchur2018pitfalls}, with a maximum of $500$ epochs.
We terminated training if the validation loss fails to improve for $100$ epochs, following the epoch termination condition outlined in \cite{shchur2018pitfalls}, with a maximum of $500$ epochs. % lee2019self
All experiments were conducted on $1 \times$ NVIDIA A100 GPU. \\
\textbf{\textit{Experimental Setup.}} 
We compared our approach against $6$ baseline methods, including $3$ static models (SVM \cite{suykens1999least}, GCN \cite{kipf2016semi} with global pooling, and DiffPool \cite{ying2018hierarchical}), and $3$ dynamic brain network embedding methods (LSTM \cite{dvornek2017identifying}, ST-GCN \cite{gadgil2020spatio}, and FE-STGNN \cite{chen2023fe}).
The $\beta$ parameter is set to $0.5$ for all experiments.
We conducted a search for optimal $\lambda$ parameter within the range of $[0.1, 0.3, 0.5, 0.7, 0.9]$ (See Fig. \ref{parameter} for details). The resulting values were $\lambda=0.3$ for HCP and $\lambda=0.5$ for OASIS.

% $6$ baselines were used for comparison, including $3$ static models (i.e., SVM \cite{suykens1999least}, GCN \cite{kipf2016semi} with a global pooling, and DiffPool\cite{ying2018hierarchical}), and $3$ dynamic graph models designed for temporal brain network embedding (e.g., LSTM \cite{dvornek2017identifying}, ST-GCN \cite{gadgil2020spatio}, and FE-STGNN \cite{chen2023fe}).
% We search the $\lambda$ parameter (details in \textcolor{blue}{Supplementary}) in the range of $[0.1, 0.3, 0.5, 0.7, 0.9]$ and determine $\lambda=0.3$ and $\lambda=0.5$ for HCP and OASIS, respectively.

\vspace{-1em}
\subsection{Brain Network Predictions}
% \vspace{-1em}
\textbf{\textit{Classification Tasks.}}
$\epsilon 4$ allele is a strong risk factor for the Alzheimers' Disease (AD) \cite{serrano2021apoe}.
Table \ref{classification} presents classification results for gender on HCP, as well as for AD and $\epsilon 4$ on OASIS.
It shows that our model achieves the highest accuracy across all tasks compared to other methods. 
Meanwhile, the comparison between results obtained with and without structural connectivity (SC) demonstrates the importance of anatomical (or spatial) constraints on effective network representation learning.
Furthermore, the dynamic methods consistently outperform the static methods, indicating their efficacy in brain network analysis by capturing brain dynamics. \\
% The classification results for gender on HCP, for AD and $\epsilon 4$ on OASIS are presented in Table \ref{classification}, which shows that our model achieves the best accuracy for all these tasks among all methods.
% We compare the results yielded by our framework with and without using structural connectivity (SC), and almost all results indicate the importance of the anatomical constraints on effective network representation learning. 
% Compared with static methods, the results of dynamic methods indicates that it is more effective in brain network analysis by modeling brain dynamics. \\
\textbf{\textit{Regression Tasks.}}
The Mini-Mental State Exam (MMSE \cite{arevalo2015mini}) serves as a quantitative assessment tool for cognitive status in adults. 
The Diagnostic and Statistical Manual of Mental Disorders (DSM \cite{american2013diagnostic}) offers a comprehensive measure system for mental disorders utilized by mental health professionals worldwide. Within the DSM system, DSM-Depr and DSM-Antis gauge two mental disorders linked to depression and rebellious personality, respectively.
Table \ref{regression} summarizes the regression results for DSM and MMSE on the HCP and OASIS datasets, showing that our model outperforms all baseline methods with lowest mean absolute values.

% The Mini-Mental State Exam (MMSE \cite{arevalo2015mini}) is a quantitative measure of cognitive status in adults.
% The Diagnostic and Statistical Manual of Mental Disorders (DSM \cite{american2013diagnostic}) is a comprehensive classification system for mental disorders used by mental health professionals in the United States and around the world, where DSM-Depr and DSM-Antis measures two mental diseases related to depression and rebellious personality, respectively. 
% The DSM and MMSE regression results on the HCP and OASIS datasets are summarized in Table \ref{regression}, which also demonstrates that our model outperforms all baselines. 

\begin{figure}[t]
\centering
\includegraphics[width=1.0\textwidth]{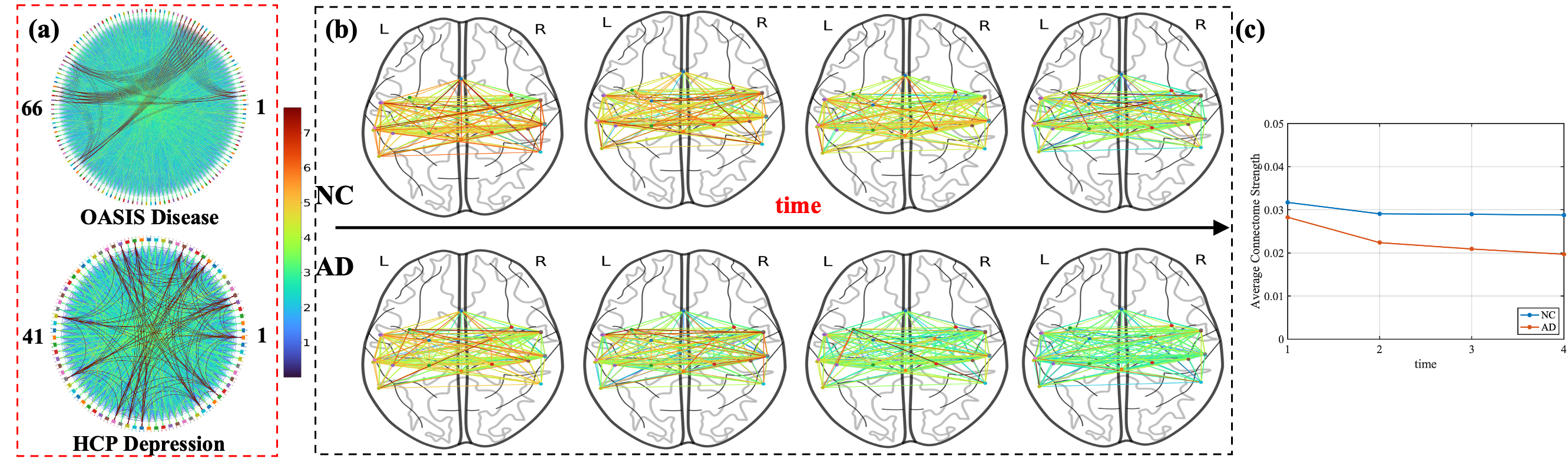}
\caption{(a) illustrates the importance of various effective connectomes (i.e., $|\gamma|$) for disease classification and DSM-Depr regression, with the most crucial connectomes highlighted in bold \textcolor{red}{red}. 
(b) visualizes the brain dynamics of the identified effective connectomes during an fMRI scan period, where colors tending towards red indicate large values.
(c) quantifies the change in the average strength of identified connectomes during an fMRI scan period.}
\label{explain}
\vspace{-1em}
\end{figure}

\subsection{Biological Insights and Model Interpretability}
We provided two distinct biological insights from our interpretable framework. 
Firstly, we utilized the designed parameter ($\gamma$) to identify the most crucial effective connectomes for various prediction tasks. Specifically, we pinpointed the top $400$ and $256$ connectomes (highlighted in bold \textcolor{red}{red} curve in Fig. \ref{explain}(a)) with the highest $|\gamma|$ weights for disease classification on OASIS and DSM-Depr regression tasks, respectively.
Our disease classification results indicate that the highlighted connectomes are predominantly linked to the most relevant brain nodes of Alzheimer's Disease (AD), such as the right/left insula cortex, anterior/posterior cingulate gyrus, and anterior/posterior divisions of the parahippocampal gyrus. 
Additionally, connectomes associated with AD-relevant subnetworks, such as the Default Mode Network (DMN) \cite{dennis2014functional,yildirim2019default}, are highlighted. Similarly, connectomes connected to the most relevant brain nodes (e.g., left/right amygdala, hippocampus and orbitofrontal) of depression are identified from DSM-Depr regression. 
The Salience Network (SN) subnetwork, crucial for emotional regulation \cite{pinto2023emotion}, is also highlighted.
Furthermore, we present the brain temporal dynamics of the identified connectomes in Fig. \ref{explain}(b), visualizing the related $\gamma \odot A^{f}$ derived from the disease classification task at each of the four time-points to illustrate how the effective connectomes change during an fMRI scan period. To quantify this change, we show the average of these $\gamma$ weighted connectomes in Fig. \ref{explain}(c). It demonstrates that the causal influence strength of the normal control(NC) group and the AD group decays simultaneously over time. However, the degree of decline in the AD group is more pronounced than in the NC group.
\vspace{-1em}

\section{Conclusion}
% We introduce an interpretable spatio-temporal framework for learning representations of brain effective networks, employing directed graph embedding layers and modeling brain dynamics via ordinary differential equations. 
% The experimental results show our framework outperforms existing methods in various clinical phenotype predictions, identifying the connectomes associated with various clinical phenotypes and illustrating the dynamic causal influence strengths throughout fMRI scan periods. Future work will explore dynamic causal influences at the level of brain ROIs.

We propose an interpretable spatio-temporal framework with directed graph embedding layers for learning brain effective network representations, leveraging ordinary differential equations to model brain dynamics.
Our framework contributes to important clinical prediction tasks, pinpointing important connectomes linked to different clinical phenotypes and illustrating dynamic causal influence strengths across fMRI scan periods.
Future work will investigate dynamic causal influences at the level of brain ROIs.

\section*{Acknowledgments}
This study was partially supported by the Presidential Research Fellowship (PRF) in the Department of Computer Science at the University of Texas Rio Grande Valley, and the UTRGV seed grant, as well as by the NSF (2112631, 2045848, 2319449, 2319450, 2319451), and the NIH (R01AG071243, R01MH125928, U01AG068057). \\
This work used Bridges-2 at Pittsburgh Supercomputing Center from the Advanced Cyberinfrastructure Coordination Ecosystem: Services \& Support (ACCESS) program, which is supported by National Science Foundation grants \#2138259, \#2138286, \#2138307, \#2137603, and \#2138296.
Data were provided in part by the Human Connectome Project, WU-Minn Consortium (Principal Investigators: David Van Essen and Kamil Ugurbil; 1U54MH091657) funded by the 16 NIH Institutes and Centers that support the NIH Blueprint for Neuroscience Research; and by the McDonnell Center for Systems Neuroscience at Washington University; and were provided in part by OASIS-3.

\bibliographystyle{splncs04}
\bibliography{ref}
% \bibliography{ref}

\end{document}